\documentclass[letterpaper, 10 pt, conference]{ieeeconf}

\IEEEoverridecommandlockouts          

\usepackage{graphics} %
\usepackage{epsfig} %
\usepackage{mathptmx} %
\usepackage{times} %
\usepackage{amsmath} %
\usepackage{amssymb}  %
\usepackage{xcolor}
\usepackage{url}
\usepackage[noadjust]{cite}

\usepackage{multicol}
\usepackage{graphicx}
\usepackage{caption}
\usepackage{times}
\usepackage{float}
\usepackage{listings}
\usepackage{mdframed}
\usepackage{soul}
\usepackage{inconsolata}
\usepackage{fancyvrb}
\usepackage[font=small]{caption}
\usepackage{cleveref}
\let\labelindent\relax
\usepackage{enumitem}
\usepackage{booktabs}
\usepackage{adjustbox}
\usepackage{pifont}
\usepackage{balance}
\usepackage{subcaption}
\usepackage[symbol]{footmisc}
\usepackage{tikz,lipsum}
\usepackage{soul}
\usepackage[most]{tcolorbox}
\usepackage{multirow}
\usepackage{array}
\usepackage{makecell}

\renewcommand{\baselinestretch}{0.97}

\newcommand{\greenyesmark}{\textcolor{green}{\ding{51}}} %
\newcommand{\rednomark}{\textcolor{red}{\ding{55}}}   %

\newcommand{\rev}[1]{\textcolor{black}{#1}}
\newcommand{\reva}[1]{\textcolor{black}{#1}}

\newcommand{\datasetname}{OpenRoboCare}

\title{
\textbf{\datasetname}: A \rev{Multimodal} Multi-Task Expert Demonstration Dataset for Robot Caregiving
\vspace{-0.2cm}
}

\author{$^1$Xiaoyu Liang, $^1$Ziang Liu, $^3$Kelvin Lin$^*$, $^1$Edward Gu$^*$, $^1$Ruolin Ye, $^4$Tam Nguyen \\$^2$Cynthia Hsu, $^1$Zhanxin Wu, $^1$Xiaoman Yang,  $^1$Christy Sum Yu Cheung \\$^3$Harold Soh, $^2$Katherine Dimitropoulou, $^1$Tapomayukh Bhattacharjee}

\begin{document}

\twocolumn[{%
\renewcommand\twocolumn[1][]{#1}%

\maketitle
\thispagestyle{empty}
\pagestyle{empty}
\begin{center}
    \begin{minipage}{1.0\textwidth}
        \centering
        \vspace{-0.8cm}
        \includegraphics[width=\textwidth, trim={5pt 5pt 5pt 0pt}, clip]{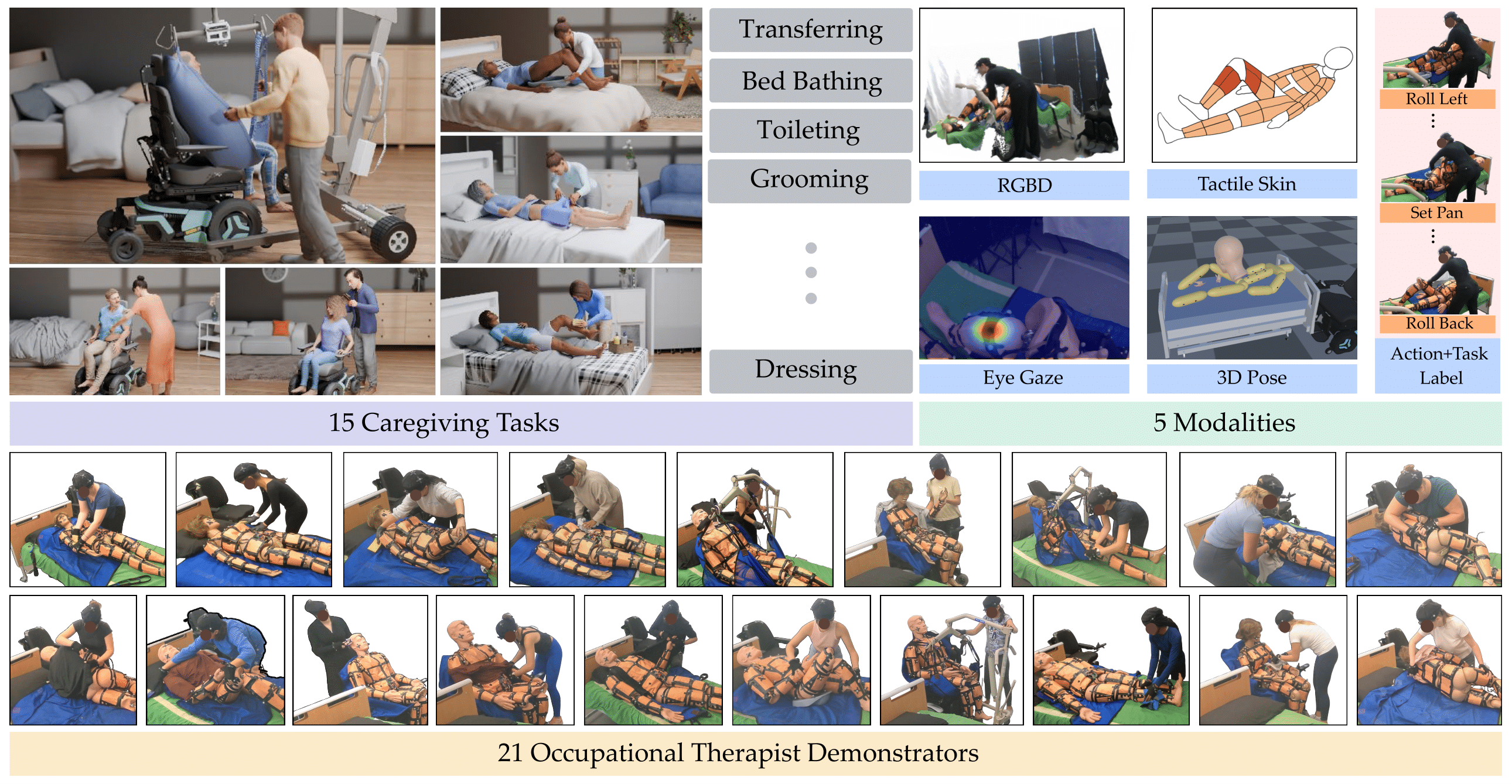}
        \captionof{figure}{\small Overview of OpenRoboCare dataset for robot caregiving, featuring 21 occupational therapists demonstrating 15 common caregiving tasks, captured across 5 data modalities. It consists of 315 sessions, totaling 19.8 hours, with a collection of 31,185 samples.}
        \label{fig:teaser}
        \vspace{-0.2cm}
    \end{minipage}
\end{center}
}]

\footnotetext[0]{$^1$\textit{Cornell University}, Ithaca, NY, USA. $^2$\textit{Columbia University}, New York City, NY, USA. $^3$\textit{National University of Singapore}, Singapore. $^4$\textit{University of Massachusetts Lowell}, Lowell, MA, USA}
\footnotetext[2]{\scriptsize This work was partly funded by NSF IIS \#2132846, CAREER \#2238792,  and a Cornell–NUS Global Strategic Collaboration Award. We thank Yapeng Teng for data labeling tool setup. We thank Tasbolat Taunyazov, Crystal Liu, Moustafa Kassem, Xia Yan Zhao, Shannon Liu, Luke Kulm, Tasmin Sangha, Mandy Chen, Gavin Chen, Nancy Davila Belendez, Max Tweedale, Allyson Daos, Emily Avisado for their efforts in data processing and labeling.}

\begin{abstract}
We present \datasetname{}, a \rev{multimodal} dataset for robot caregiving, capturing expert occupational therapist demonstrations of Activities of Daily Living (ADLs).
Caregiving tasks involve complex physical human-robot interactions, requiring precise perception under occlusions, safe physical contact, and long-horizon planning.
While recent advances in robot learning from demonstrations have shown promise, there is a lack of a large-scale, diverse, and expert-driven dataset that captures real-world caregiving routines.
To address this gap, we \rev{collect} data from 21 occupational therapists performing 15 ADL tasks on two manikins. The dataset spans five modalities---\rev{RGB-D} video, pose tracking, eye-gaze tracking, task and action annotations, and tactile sensing, providing rich \rev{multimodal} insights into caregiver movement, attention, force application, and task execution strategies.
We further analyze expert caregiving principles and strategies, offering insights to improve robot efficiency and task feasibility.
Additionally, our evaluations demonstrate that \datasetname{} presents challenges for state-of-the-art robot perception and human activity recognition methods, both critical for developing safe and adaptive assistive robots, highlighting the value of our contribution. See our website for additional visualizations: \rev{\url{https://emprise.cs.cornell.edu/robo-care/}}.

\end{abstract}

\vspace{-0.2cm}
\section{Introduction}

\begin{table*}[h]
    \centering
    \caption{\small Comparison of \datasetname{} with existing datasets.}
    \renewcommand{\arraystretch}{1.2}
    \begin{tabular}{l@{\hskip 4pt} c@{\hskip 4pt} c c@{\hskip 8pt} c@{\hskip 8pt} c c c c@{\hskip 4pt} c c c c c}
        \toprule
        & \multicolumn{3}{c}{\textbf{Context}} & \multicolumn{4}{c}{\textbf{Statistics}} & \multicolumn{6}{c}{\textbf{Data Modality}} \\  
        \cmidrule(lr){2-4} \cmidrule(lr){5-8} \cmidrule(lr){9-14}
        \textbf{Dataset} & Care.$^1$ & Pop.$^2$ & Demo.$^3$ & \#Subj.$^4$ & \shortstack{\#Tasks \\ $_{(ADLs)}$} & \#Hrs & Samples$^5$ & IMU & Video & \shortstack{Pose \\ $_{CG\mid CR}$}$^6$ & Gaze & Tactile & Act.$^7$ \\
        \midrule
        Bagewadi et al.~\cite{human_robot_hugging} & \rednomark & TD & User & 33 & 1(0) & 2.94 & 485 & \greenyesmark & RGB &  M \scriptsize{(-$\mid$\greenyesmark)} & \rednomark & Partial & A,T \\
        SBU Kinect~\cite{sbu_kinect} & \rednomark & TD & User & 7 & 8(0) & - & - & \rednomark & RGB-D & V \scriptsize{(-$\mid$\greenyesmark)} & \rednomark & \rednomark & T\\
        TacAct~\cite{tacact} & \rednomark & TD & - & 50 & 12(0) & - & - & \rednomark & \rednomark & \rednomark & \rednomark & Partial & A\\
        SONAR~\cite{sonar} & \greenyesmark & E & Expert & 14 & 23(5) & 37.3\footnote[1]{} & 36006 & \greenyesmark & \rednomark & M \scriptsize{(\greenyesmark$\mid$\rednomark)} & \rednomark & \rednomark & T\\
        Kaczmarek et al.~\cite{10.1145/3558884.3558891} & \greenyesmark & ML & Expert & 7 & 9(1) & 0.7 & 132 & \greenyesmark & RGB & \rednomark$\mid$\rednomark & \rednomark & \rednomark & T\\
        HARMONIC \cite{harmonic} & \greenyesmark & TD & User & 24 & 1(1) & 5 & 600 & \greenyesmark & RGB & V \scriptsize{(-$\mid$\greenyesmark)} & \greenyesmark & \rednomark & T\\
        \midrule
        \textbf{\datasetname{}} \textbf{(ours)} & \greenyesmark & SML & Expert & 21 & 15(5) & 19.8 & 31185 & \rednomark & RGB-D & M,V \scriptsize{(\greenyesmark$\mid$\greenyesmark)} & \greenyesmark & Full & A,T\\
        \bottomrule
    \end{tabular}
    \\
    \vspace{2pt}
    \scriptsize
    $^1$Caregiving. $^2$Target population. (TD: typical developing. E: elderly. ML: mobility limitation. SML: severe mobility limitation.) $^3$Demonstration type. $^4$Number of subjects. $^5$Total number of samples (modalities $\times$ tasks $\times$ subjects $\times$ time in hours)
    $^6$Pose data. (M: motion capture, includes both optical and inertial. V: pose calculated from videos. CG: caregiver pose. CR: care recipient or user pose.)
    $^7$Action annotation. (T: task-level. A: action-level.) \footnote[1]{}Re-calculated after removing non-documented activities.
    
    \label{tab:comparison}
    \vspace{-0.5cm}
\end{table*}

According to the World Health Organization~\cite{who2022}, approximately 1.3 billion people live with significant physical limitations, many of whom require assistance with Activities of Daily Living (ADLs)~\cite{adl}.
However, the demand for qualified caregivers and therapists far exceeds the number of available trained professionals~\cite{scales_caregiver_training}. Assistive robotics \rev{has potential} to support the caregiving process and address these issues to some extent. Recent work in \emph{robot caregiving} in feeding\rev{~\cite{feeding1,feeding2,10660886,jenamani2025feast}}, dressing~\cite{dressing1,dressing2}, bathing~\cite{bathing1,bathing2}, and transferring~\cite{transferring1} among other ADLs presents tremendous promise.

Despite these advancements, robot caregiving still faces significant technical challenges.
For example, consider the requirements for assisted bed bathing~\cite{bathing1,bathing2}:
accurate perception of human state, often under occlusions; bimanual and mobile manipulation of human limbs with critical safety constraints; long-horizon planning under uncertainty; \rev{personalization to users’ physical function~\cite{liu2025grace} and preferences~\cite{silver2025coloring}; and adaptation in response to human feedback}.
\\
\indent To address these challenges, recent work has considered learning-based approaches for specific caregiving tasks~\cite{sundaresan2022learning,ha2024repeat,ye2025cart}.
These prior works \rev{collected} their own datasets, typically in simulation, that are tailored to the task of interest.
This \emph{task-specific} data collection is in contrast to recent advances in general robot learning where large diverse datasets are used to train foundation models~\cite{khazatsky2024droid,o2024open}.
Existing caregiving datasets are also limited in \emph{data modalities}, often featuring vision~\cite{ye2025cart} or haptics~\cite{10.1145/3558884.3558891} but not both.
Furthermore, with a few notable exceptions~\cite{10.1145/3558884.3558891,sonar}, existing caregiving datasets do not feature data collected from \emph{expert} human caregivers or occupational therapists.
These datasets are therefore lacking the extensive practical knowledge accumulated by experts through years of experience and training.\\
\indent To bridge this gap, we present \textbf{\datasetname{}} (Fig.~\ref{fig:teaser}), the first \emph{multi-task, \rev{multimodal}, and expert-collected} dataset for robot caregiving.
\datasetname{} features \rev{expert demonstrations from 21 occupational therapists (OTs)} in 15 distinct ADLs with data captured from five modalities: \rev{RGB-D} video, tactile sensing, pose tracking, eye-gaze tracking, and action annotations.
We \reva{provide} tactile sensing with a custom whole-body piezo-resistive skin, pose tracking with motion capture, eye-tracking with Pupil Labs glasses, and action labeling via natural language during data collection.
The OTs \rev{demonstrated} their assistance on two manikins with different genders and \reva{body} weights.
Our data collection protocol \rev{was} designed in collaboration with an expert occupational therapist and co-author, ensuring that the task procedures and setup closely resemble real-world caregiving scenarios.\\
\indent \reva{In addition to releasing \datasetname{} as an \emph{open-source dataset}, we conduct a comprehensive quantitative analysis to inform task execution and physical interactions with care recipients in robot caregiving. In collaboration with an expert OT, we distill guiding principles in OT practice and identify specific techniques that exemplify these principles across tasks.
We present \datasetname{} as a benchmark for robot perception and human activity recognition in caregiving scenarios. While state-of-the-art methods perform poorly out of the box, fine-tuning on a small subset of \datasetname{} leads to significant performance gains.
}
Overall, our findings highlight the richness and complexity of expert caregiving strategies, positioning \datasetname{} as a critical resource for advancing multimodal learning in robot caregiving.

\vspace{-0.2cm}
\section{Related Work}

\textbf{Caregiving Datasets}
Previous works in rehabilitation and public health collected survey and interview-based data on caregiving, e.g., for older adults~\cite{AARPandNationalAllianceforCaregiving2020,goodwin2024leveraging,hyejin2021adl} and those with spinal cord injuries~\cite{wilson2022systematic}.
These efforts focus on the health, social, and financial aspects of caregiving, rather than the physical caregiving process.
Works close to ours \rev{either} \rev{did} not collect data from expert caregivers \cite{harmonic}, or \rev{lacked} multimodality \cite{sonar, 10.1145/3558884.3558891}.
\datasetname{} is the first dataset for multi-task, \rev{multimodal}, expert caregiving (\rev{Table~\ref{tab:comparison}}).

\textbf{Physical HHI \& HRI Datasets}
Our work is also related to previous efforts in human-human and human-robot interaction \rev{that collected} \rev{multimodal} and multi-task data, e.g., for activity recognition~\cite{industrial_phri,tactile_pinch_forearm}.
For example, Bagewadi et al. \cite{human_robot_hugging} collected data of human-robot hugging interactions using wearable sensors.
The SBU Kinect Interaction~\cite{sbu_kinect} human-human dataset similarly considered hugging among other physical activities such as kicking and punching.
TacAct~\cite{tacact} collected high-fidelity tactile data of a human touching a robot arm.
We similarly collect high-fidelity tactile data, but \rev{placed} sensors on the human body rather than the robot arm.

\textbf{Human Activities Datasets}
Our work is also more broadly related to the literature on human activity recognition~\cite{saleem2023toward}.
Most related are datasets for activity recognition that are situated in homes and hospitals~\cite{simadl,pmlr-v205-patel23a,Tonkin_2023,sanchez2008activity}.
In experiments, we evaluate a state-of-the-art method~\cite{yang2023vidchapters} for activity recognition on \datasetname{} and show that our tasks present significant challenges and can drive further progress in the field.

\vspace{-0.2cm}
\section{Task Selection and Caregiving Protocol}

Our goal is to collect a \rev{multimodal}, multi-task, expert-driven dataset that facilitates robot caregiving research.
In this section, we describe the \rev{included tasks} and the protocol for expert caregivers completing the tasks.
This data collection protocol \rev{was} approved by \rev{the} Cornell IRB.
\vspace{-0.2cm}
\subsection{OT-in-the-loop Protocol Design}
We \rev{selected tasks and designed} our data collection in close consultation with an expert OT collaborator who also has extensive experience in OT education.
Our setup \rev{follows} standard clinical guidelines for training OTs~\cite{stein2024pocket,bell2021activities}.
In designing the setup, we especially consider individuals with quadriplegia, a significant sensorimotor impairment that results in a lack of control and movement of the upper limbs, trunk, lower limbs, and pelvic organs \rev{and requires complete assistance with basic ADLs} \cite{rybski2024rehabilitation}.

\vspace{-0.2cm}
\subsection{Task Selection}
We consider five of the six basic \rev{A}ctivities of \rev{D}aily \rev{L}iving (ADLs):
bathing, toileting, dressing, transferring, and grooming.
Feeding is excluded as it requires substantially different caregiving skills and has been extensively studied~\cite{NewmanHARMONIC2021,DVN/8TTXZ7_2018}. 
Within the five basic ADLs, we consider 15 tasks (Fig.~\ref{fig:setup}): one each from bathing, toileting, and grooming; two from transferring; and 10 from dressing \rev{capturing diverse garment types and scenarios}.
We develop caregiving protocols that mirror real-life routines of care recipients with quadriplegia.
We next describe the ADLs and tasks in detail\rev{, highlighting key aspects relevant to robot caregiving}.\\
\indent \textbf{Bathing}
Caregivers perform a full-body sponge bath on a manikin lying on a hospital bed.
They are instructed to pat the skin gently, as they would with a care recipient to minimize the risk of discomfort or injury. Key aspects of interest include the amount of force applied, techniques for cleaning hard-to-reach areas, and strategies for adjusting the manikin's position to access its back.\\
\indent \textbf{Toileting}
Caregivers assist a manikin with toileting \rev{using a bedpan} while it is lying on a hospital bed, a common method for individuals with limited mobility or a high risk of injury that prevents the use of a regular toilet. The task requires caregivers to lift the manikin’s hips to position the bedpan underneath and subsequently remove it for emptying. Key aspects of interest include the techniques used to lift and stabilize the manikin’s hips when handling manikins of varying weights, as well as the hand placements and stabilization strategies employed.\\
\indent \textbf{Dressing}
Caregivers dress or undress a manikin.
Dressing requires different strategy sequences and safety considerations for different body segments (upper/lower body), garment types, supporting surfaces (bed/wheelchair), and body position (lying down/sitting). 
To capture these variations, caregivers dress and undress the manikin in t-shirts, vests, and shorts while the manikin is lying on a bed or sitting in a wheelchair.
In total, we define 10 tasks (see Fig.~\ref{fig:setup}). Key aspects of interest include the ways caregivers prepare clothing, coordinate bimanual movements, and handle care recipients' body parts.\\
\indent\textbf{Transferring}
Caregivers transfer a manikin between a bed and a wheelchair using a Hoyer sling with a mechanical lift (see Fig.~\ref{fig:setup}).
We consider bed-to-wheelchair and wheelchair-to-bed transferring as two separate tasks.
Key aspects of interest include the ways in which caregivers use the Hoyer sling, secure the sling properly, operate the lift, guide the manikin’s position during transfer, and safely release the manikin after the transfer.\\
\indent \textbf{Grooming}
Caregivers brush the manikin’s hair while it is seated in a wheelchair.
\rev{Key aspects of interest include caregiver hand coordination, posture adjustments, and control strategies during the grooming procedure.}

\vspace{-0.2cm}
\subsection{Caregiver Protocol}
\rev{Data collection spanned two weeks and involved 21 OTs.}
Each participant performed 15 tasks on a male or female manikin, one trial per task, taking approximately one hour total.
Before data collection, participants completed IRB consent forms and demographic surveys. Upon arrival, they received a detailed briefing covering the study objectives, task instructions, equipment usage, and data collection procedures. Each caregiver was then fitted with motion capture gloves, a motion capture hat, and eye-tracking glasses. We calibrated the motion capture system and the eye-tracking glasses with an OT before each data collection session.

\section{Data Collection Setup}
\label{sec:data-collection}

\begin{figure*}[ht]
  \centering
    \includegraphics[width=\textwidth, trim={35pt 30pt 25pt 15pt}, clip]{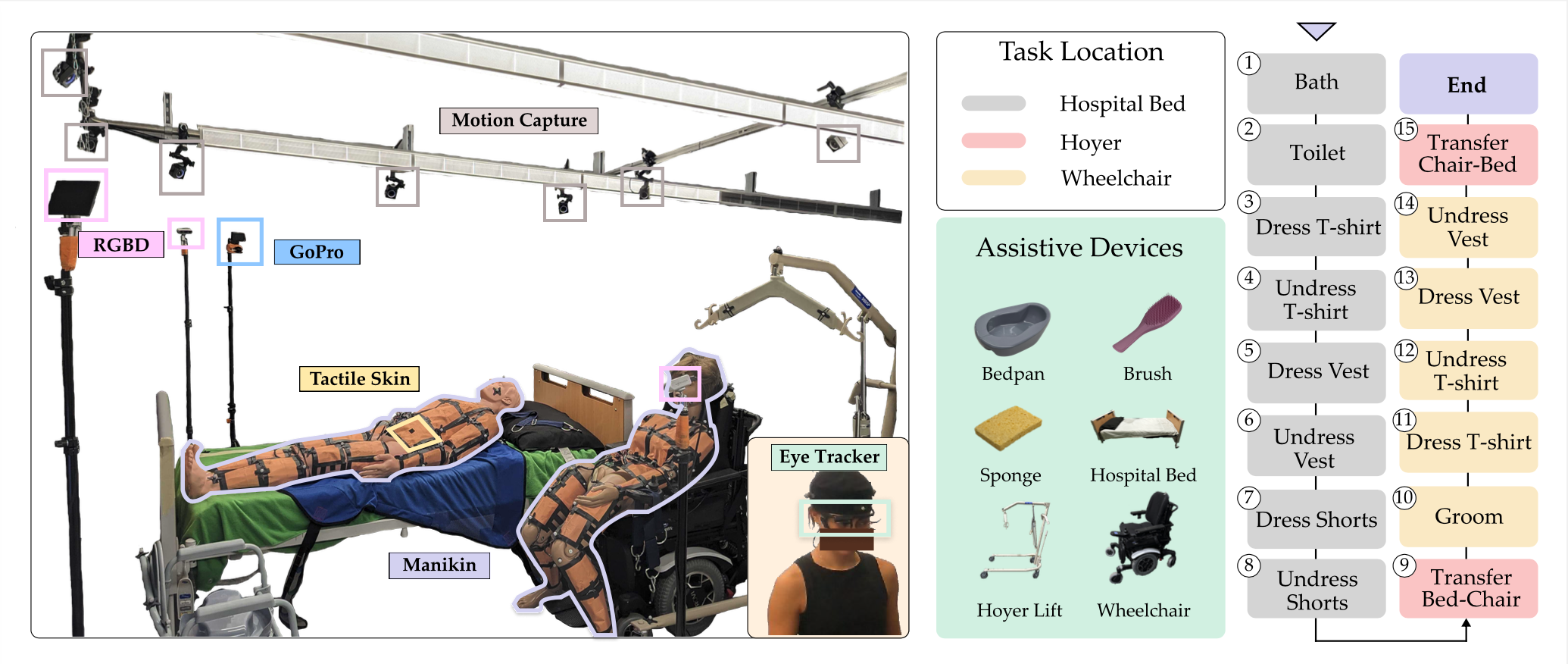}
    \captionof{figure}{\small Data collection setup and procedure. \textbf{Left}: setup of sensors and equipment. \textbf{Center}: assistive devices used by caregivers. \textbf{Right}: sequence of tasks performed by each caregiver.}
    \label{fig:setup}
    \vspace{-0.5cm}
\end{figure*}

In this section, we describe the environment and data recording setup. Data collection took place in an enclosed space measuring $3.68\text{m} \times 3.68\text{m}$, designed to simulate a realistic in-home caregiving setting. The hospital bed, wheelchair, and Hoyer sling \rev{were reset to the same initial position} throughout the sessions.
See Fig.~\ref{fig:setup} for an overview.

\vspace{-0.2cm}
\subsection{Hospital Manikins}
In consultation with our OT expert, we select two hospital manikins. One male (Rescue Randy, 150 lbs, 6 ft 1 in) and one female (Simple \rev{Susie}, 37.26 lbs, 5 ft 5 in), both with anthropomorphic \rev{dimensions} and joints.
Manikins are frequently used in OT clinical training to simulate real-life conditions \cite{mortimer2018comparison}.
\rev{They allow us to standardize caregiving tasks and collect reliable, repeatable data without \reva{inconveniencing} real patients. While manikins lack partial agency or resistance, they effectively represent passive full-assistance scenarios that are common in OT practice. The strategies demonstrated by expert caregivers offer valuable insights for adapting to interactions with partially active patients.}

\vspace{-0.2cm}
\subsection{Assistive Devices}
\rev{The data collection setup involves various assistive devices} to replicate realistic caregiving environments. The hospital bed (Invacare ETUDE HC Hi-Lo) and electric wheelchair (ROVI X3) are positioned next to each other in a fixed arrangement. A Hoyer sling (Invacare 9805P) is also included for transferring tasks. For specific caregiving tasks, assistive devices are provided: \rev{a bathing sponge for bathing, a bedpan for toileting, and a brush for grooming.}
\vspace{-0.2cm}
\subsection{Sensing Modalities}
\label{sec:sensing_modalities}
\textbf{RGB-D Videos}
We use three Intel RealSense D435i RGB-D cameras positioned around the scene to capture visual and geometric data of the caregiver’s movements, interactions with the manikin and assistive devices, and the resulting manikin motion. Two cameras are placed at different angles facing the hospital bed, while a third camera is positioned behind the bed facing the wheelchair. 

\begin{figure}[!t]
  \includegraphics[width=\linewidth, trim={15pt 12pt 5pt 10pt}, clip]{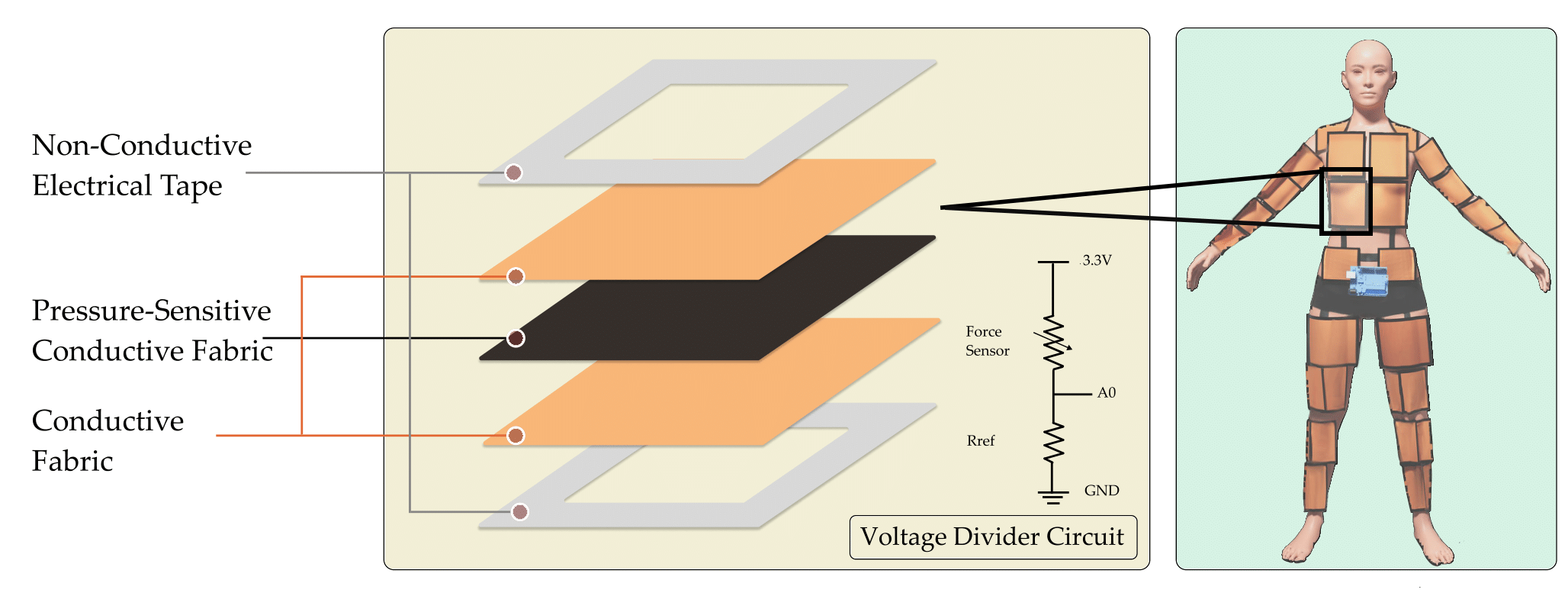}
        \captionof{figure}{\small Tactile skin design and layout of sensors on manikin.}
        \label{fig:Tactile Skin Design}
        \vspace{-0.8cm}
\end{figure}

\textbf{Tactile Skin}
We develop a custom tactile skin to fit the manikins and record physical interactions between the caregiver and the manikin. The sensor design is guided by three key considerations: (1) \textit{customizability} to accommodate various manikin body shapes and sizes, (2) \textit{flexibility} to ensure secure attachment to curved surfaces, and (3) \textit{durability} to withstand pressure exerted by the manikin’s weight.

\textit{Hardware Design}:
We design resistive tactile sensors using specialized fabrics that are lightweight, flexible, and durable (Fig.~\ref{fig:Tactile Skin Design}). Each sensor consists of a pressure-sensitive Velostat layer, sandwiched between two copper conductive fabric layers and secured with non-conductive electrical tape. The sensor's resistance decreases as force is applied, enabling force measurement.
A voltage divider circuit converts resistance changes into analog voltage signals, which are processed by an Arduino Uno. \rev{The sensor exhibits a nonlinear but stable voltage response across the 0.05 to 3 N/cm² pressure range. It has low hysteresis observed at forces below 5N and a maximum hysteresis error of $7.09\%$ at higher loads. See website~\cite{website} for details.}

\textit{Sensor Placement}:
A total of 88 resistive sensors are developed, with 44 sensors placed on each manikin (Fig.~\ref{fig:Tactile Skin Design}). The sensors are evenly distributed across the manikin's body: 7 on each arm, 8 on each leg, and 14 across the front and back \rev{of the torso}, aiming to maximize coverage. Each sensor covers an average area of 50 square inches\rev{, with gaps between adjacent sensors kept under 1 mm}.

\textit{Calibration and Processing}:
Prior to data collection, each tactile sensor is calibrated using an ATI \rev{f}orce/\rev{t}orque (F/T) sensor \reva{to accurately map voltage readings} to force values. Before each task, the tactile skin is tared to eliminate any baseline offset, ensuring consistent force measurements.

\textbf{Pose Tracking}
We use a motion capture system equipped with 12 OptiTrack \rev{PrimeX 13} cameras to track the movements of both the manikin and the caregiver. The caregiver wears a hat and gloves with motion capture markers to accurately track hand and head movements.
For manikin pose tracking, we employ rigid body marker sets to define each body segment.

\rev{Occlusions pose a fundamental challenge in real-world caregiving and are critical for robotic systems to overcome. In our dataset}, occlusions caused by clothing in dressing tasks and slings in transferring tasks often led to tracking failures for both the manikin’s pose and the caregiver’s hand positions. \rev{To address this, we manually labeled body keypoints for a subset of the dataset using RGB images from three calibrated cameras, originally used for RGB-D video capture (details in Section \ref{sec:sensing_modalities}). These annotations were then used to train a YOLOv11~\cite{Jocher_Ultralytics_YOLO_2023} pose detector, which automatically labeled the remaining data with \reva{little} human supervision. We estimated 3D positions by averaging triangulated results from all camera pairs. While limited to three views, this approach provides a practical and scalable solution for occlusion handling.}

\textbf{Eye Tracking}
To analyze caregiver visual attention during tasks, we equip participants with Pupil Labs eye-tracking glasses to capture first-person video and 2D gaze data. We use 3D pose tracking of the caregiver's head as a proxy for the eye-tracking glasses' pose. During post-processing, \rev{we apply a low-pass filter to smooth the gaze data}. \rev{We correct minor shifts in the glasses by re-aligning the gaze vector}.

\begin{figure*}[ht]
  \centering
    \includegraphics[width=\textwidth]{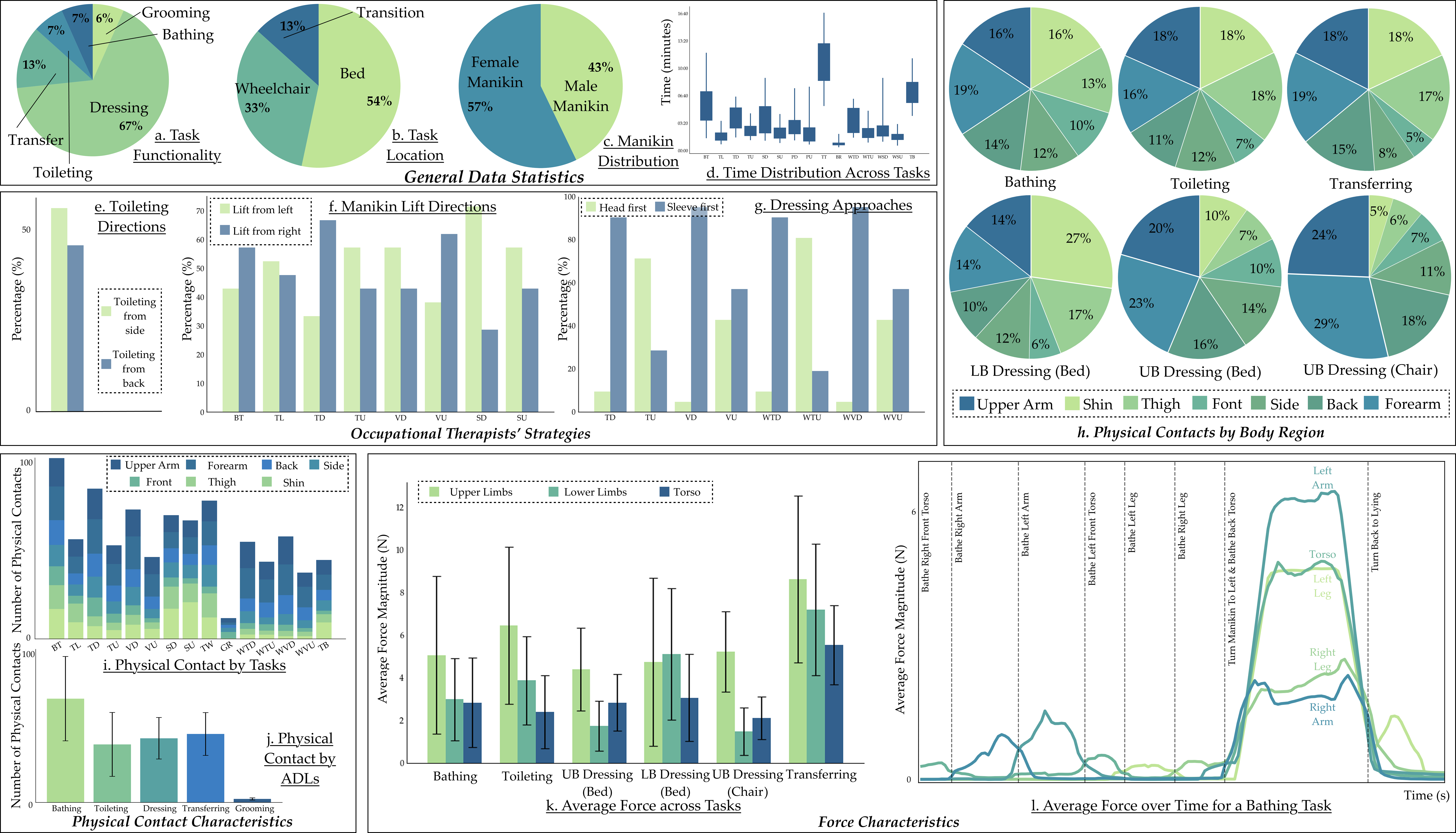}
    \raggedright
    \tiny
    BT (Bathing), TL (Toileting), TD (Dressing T-shirt), TU (Undressing T-shirt), VD (Dressing Vest), VU (Undressing Vest), SD (Dressing Shorts), SU (Undressing Shorts), WTD (Wheelchair Dressing T-shirt), WTU (Wheelchair Undressing T-shirt), WVD (Wheelchair Dressing Vest), WVU (Wheelchair Undressing Vest), TW (Transfer to Wheelchair), TB (Transfer to Bed), and GR (Grooming)

    \captionof{figure}{\small Analysis of Dataset Characteristics. We analyze the diversity of the dataset across different aspects: (a-c) general data collection statistics; (d-f) occupational therapists' strategies; (g) time duration across tasks; (h-j) physical contact characteristics; and (k-l) force magnitude information.} %
    \label{fig:stats}
    \vspace{-0.5cm}
\end{figure*}

\textbf{Task and Action Labeling}
ADL tasks like Hoyer sling transfers are long-horizon activities with multiple steps. Understanding caregiver actions at each stage enables segmentation into modular components. To support this, we record video and audio using a GoPro, \rev{with caregivers verbally describing their actions as they interact with the manikin}. OTs then annotate the recordings, segmenting tasks into meaningful sub-tasks based on their expertise, providing insights into task decomposition and procedural flow.

\vspace{-0.2cm}
\subsection{Sensor Synchronization} 
Due to hardware limitations, each sensor operates at a different sampling rate: RGB-D cameras at 15 Hz, tactile skin at 60 Hz, motion capture at 150 Hz, and the eye tracker at 120 Hz. \rev{To achieve temporal synchronization, all computers are synchronized with the NIST Internet Time Servers using the \texttt{chrony} service on Ubuntu and \texttt{w32tm} on Windows. For data alignment, we use the RGB-D stream (15 Hz) as the reference timeline, and extract the closest timestamped samples from the other modalities for each RGB frame to produce synchronized frames at 15 Hz.}

\section{Dataset Characteristics and Analysis}

\datasetname{} contains 315 sessions of caregiver expert demonstrations, totaling 19.8 hours of \rev{multimodal} data across 5 modalities, collected by 21 occupational therapists. The task selection covers 5 out of 6 basic activities of daily living, with 15 task variations, for a total of 31,185 expert demonstration data samples. The dataset\rev{, usage documentation, and the fine-tuned pose estimation model} is publicly available on our website~\cite{website}. In this section, we present the characteristics and unique insights within the dataset that can inform caregiving robot design.

\vspace{-0.2cm}
\subsection{Caregiver Demographics}
We recruited 21 occupational therapists (OTs), including 19 final-year OT students and 2 licensed OT professionals. The most experienced OT has over 40 years of clinical experience. 
All participants are female, ages 22 \rev{to} 75.
Collectively, they have experience working with populations with neurological conditions, stroke, traumatic brain injury, spinal cord injury, muscular dystrophy, and cerebral palsy.

\vspace{-0.2cm}
\subsection{Guiding Principles for Caregiving}
Throughout data collection, we observe various principles that OTs use to perform tasks \rev{efficiently and with minimal physical effort}.
We collaborate with an experienced OT to analyze our observations and distill underlying principles that can guide robot design for caregiving tasks. 

 \textbf{\sethlcolor{cyan!15}\hl{Principle 1} - Pre-positioning (P1)}: OTs prioritize safety by carefully preparing the care recipient before initiating a task. They ensure that the care recipient's posture, stability, joint angles, and supporting surfaces are appropriate for task execution. For example, before rolling the care recipient to the side, OTs align the trunk on the supporting surface to prevent unexpected limb trapping or unintended \rev{shifts in momentum due to} weight redistribution. %

\textbf{\sethlcolor{cyan!15}\hl{Principle 2} - Anticipation (P2)}: OTs anticipate and \rev{position} their body mechanics to support the entire task sequence, particularly for large-scale movements. They anticipate both the final position and the trajectory of the care recipient’s body and limbs, which influences task execution decisions. For example, when rolling a care recipient on the bed, an OT may position their hands on the opposite side of the body before initiating the roll. Although this places the OT \rev{at} a biomechanical disadvantage initially, it provides better control, allows for monitoring discomfort, and ensures the care recipient is positioned well for the next step. %

\textbf{\sethlcolor{cyan!15}\hl{Principle 3} - Efficiency (P3)}: OTs prioritize accuracy and timely completion of tasks to ensure efficiency. Care recipients with severe mobility limitations often have medical conditions, making efficient ADL execution crucial. Delays can lead to bradyarrhythmias, hypotension, or dizziness, while improperly placed garments may cause pressure ulcers. \rev{For example, during transfers with a Hoyer sling, OTs minimize the \reva{duration} the care recipient is lifted to reduce discomfort and physiological stress.}

\vspace{-0.2cm}
\subsection{Illustrative Caregiving Techniques}
\reva{We connect the principles to concrete techniques observed in \datasetname{}, illustrating how expert OTs ground these high-level strategies in real caregiving scenarios.}

\textbf{\sethlcolor{orange!20}\hl{Technique 1} -} \textbf{Bridge Strategy}: bending the care recipient’s knees and applying pressure behind the knees at the top of the calf to momentarily elevate the pelvis\reva{, which involves anticipating the motion (P2)}. 
This technique, often used in bed toileting to position a bedpan, requires significant caregiver effort and is most \reva{efficient} when the care recipient has a smaller body size than the caregiver \reva{(P3)}.

\textbf{\sethlcolor{orange!20}\hl{Technique 2} -} \textbf{Segmental Roll}:  gradually turning the care recipient’s body. The OT bends the care recipient’s opposite-side knee and applies pressure on the bent knee to initiate a progressive rolling motion toward the OT \reva{(P1)}. 
The pelvis moves first, followed by the upper body, shoulders, and finally the head. This technique allows for slow and controlled movement \reva{(P2)}, making it particularly useful for bed bathing and toileting, especially for care recipients prone to dizziness. Additionally, it benefits caregivers who are significantly smaller than the care recipient, as it reduces physical strain.

\textbf{\sethlcolor{orange!20}\hl{Technique 3} -} \textbf{ Wheelchair Recline During Transfer}: reclining the wheelchair to a 45-degree backward tilt before transferring the care recipient \rev{improves} positioning of the care recipient on the chair\reva{, another example of pre-positioning (P1)}. \reva{This action also minimizes the need for post-transfer adjustments and reduces physiological stress, aligning with the efficiency principle (P3).}

\textbf{\sethlcolor{orange!20}\hl{Technique 4} -} \textbf{Stabilizing Key Points of Control}: The pelvic bone, shoulders, and head serve as the primary points of control for body movement and are essential in all ADL tasks.  
To facilitate movement, OTs place their hands on key control points---such as the \rev{left or right} scapula and pelvis---to provide \rev{input to} initiate, support, and control movement\reva{, aligning with with pre-positioning and anticipation principle (P1, P2). 
This strategy maximizes the ability of the therapist to effectively and efficiently facilitate body movements, such as bed rolling, for the care recipient (P3)}.

\vspace{-0.2cm}
\subsection{Insights from Task Execution}
The dataset captures diverse task executions, offering valuable insights for training robots in caregiving-specific parameters, strategies, and workflow optimization.

\textbf{Task Duration}:
We show the duration of each task in Fig.~\ref{fig:stats}d. Care recipient transfer from bed to wheelchair is the most time-consuming task, followed by transfer from wheelchair to bed, both taking significantly longer than any other ADL. Transferring  \rev{requires maneuvering multiple assistive devices and manipulating deformable objects and human limbs, making it \reva{a long-horizon task} that can take up to 9 minutes even for expert OTs}.

\textbf{Toileting Approach}: OTs take two distinct approaches for the toileting task (Fig. \ref{fig:stats}e): (1) rolling the manikin to the side and inserting the bedpan under the body; and (2) bending the knees and \rev{lifting the manikin's hips} to \rev{place the bedpan underneath}. OTs working with the heavier manikin chose the first technique, as it requires less physical effort. 

\textbf{Manikin Lift Side Preference}:
Fig.~\ref{fig:stats}f shows the lifting-side preferences for on-bed tasks.
We do not observe any significant trend in side preferences.
For dressing tasks, we observe that OTs lift the manikin from both sides to prevent the cloth from getting stuck underneath.
All \rev{but one OT was} right-handed. \rev{We observe no correlation between hand dominance and the preferred lifting side}. In clinical practice, OTs also follow the care recipient’s preferences for lifting \reva{in addition to} their own movement preferences. To make this process efficient, OTs use specific biomechanical techniques that do not require their maximal strength.

\textbf{Dressing Approach}: 
Fig. \ref{fig:stats}g shows the distribution between two observed dressing approaches: (1) head-first, and (2) sleeve-first. Over $90\%$ \rev{of OTs prefer to insert the sleeves first when dressing the T-shirt or vest} for both hospital bed and wheelchair dressing tasks. \rev{Over $75\%$ of OTs prefer to undress the T-shirt head-first}. \rev{Dressing requires precise limb manipulation, making the sleeve-first approach preferable for better control. In contrast, undressing involves fewer constraints and does not demand precise limb guidance, making head-first approach quicker and more intuitive.}

\vspace{-0.2cm}
\subsection{Insights from Physical Interactions} 
The dataset provides empirical insights into how human caregivers distribute force and interact with different body regions across tasks.

\textbf{Physical Contact}: Different tasks require physical interactions with different body regions (Fig.~\ref{fig:stats}h). Lower-body dressing requires significantly more contact with \rev{the} shin and thigh, whereas upper-body dressing requires more contact with \rev{the} forearm and upper arm. Transferring \rev{results} in near equal contact with body regions. Physical contact differs within ADL variations, with notable differences between transferring to a wheelchair versus a bed, dressing versus undressing, lying versus sitting, and different clothing types (Fig.~\ref{fig:stats}i). Among all ADLs, bathing has the highest number of physical contacts, while grooming has the least (Fig.~\ref{fig:stats}j).

\textbf{Force Magnitude}: Additionally, \rev{the} magnitude of force exerted on the manikin varies significantly across tasks (Fig.~\ref{fig:stats}k). Transferring requires the highest force, as the manikin must be completely lifted. In general, greater force is applied to the limbs than to the torso, as the limbs act as leverage points to turn the manikin and adjust its posture. Force magnitude also varies across body regions \rev{over time} during a bathing task (Fig.~\ref{fig:stats}l). It peaks when the caregiver bathes a specific area, while turning the manikin increases force across all regions.

\vspace{-0.2cm}
\subsection{Guidelines for Robot Caregiving}
We distill insights from occupational therapist demonstrations to guide the development of caregiving robots. By analyzing gaze information, we can determine the caregiver’s area of interest, allowing robots to identify where to act.  Additionally, \rev{the caregiver's gaze shift speed} provides an estimate of how fast the robot policy should be. \reva{Predictive gazes occur when the caregiver looks at a different body part before engaging in contact with it. Caregivers have a pre-emptive timing of around 2.02 seconds. In terms of robot policy planning, this delay indicates a possible look-ahead timing to predict the next trajectory, while the robot is still currently acting on its present task.} 
\reva{Tactile sensing indicates the range and distribution of force the robot should apply. For example, gentle tasks such as bathing typically involve light contacts around 0.1–2 N, while physically demanding tasks such as repositioning or turning the body can exceed 20–30 N. This wide range highlights the need for robots to be capable of both delicate touch and high-force interaction, requiring torque-sufficient, compliant, and backdrivable actuators, along with force sensors that offer high resolution and a broad dynamic range.
Caregivers often use whole-arm contact, such as bracing the body during a roll, which differs significantly from typical robot manipulation techniques like pick-and-place. To support these interactions, robots should incorporate distributed sensing and compliance along the entire arm, not just at the end-effector, to enable safe and effective physical contact throughout the task.}
Observing caregivers’ workspaces helps define a reasonable range for designing caregiving robot hardware. Since caregiving strategies may vary based on a caregiver’s body shape, these strategies could also change depending on the robot’s embodiment. Robots can learn to coordinate multiple assistive devices by leveraging insights from human caregiving interactions. \rev{See our website for details~\cite{website}.}

\vspace{-0.2cm}
\section{Evaluation and Open Challenges}

We evaluate the state-of-the-art perception~(Sec.~\ref{Sec:Evaluation-perception}) and planning~(Sec.~\ref{Sec:Evaluation-planning}) methods on OpenRoboCare, and discuss the open challenges to address the question: \emph{What is the performance gap in existing approaches for the robot caregiving domain?} We also highlight the potential for this dataset to help \rev{advance robot vision and planning methods}.

\subsection{Perception: Human pose estimation}~\label{Sec:Evaluation-perception}
We evaluate SOTA pose detection methods on RGB images for tracking the manikin's pose. For 2D pose estimation (Table~\ref{tab:2dpose_detection}), \rev{we use mAP$_{50\text{--}95}$}, and for 3D pose estimation (Table~\ref{tab:3dpose_detection}), we use Mean Per Joint Position Error (MPJPE).

\rev{For 2D pose estimation, the off-the-shelf YOLOv11~\cite{Jocher_Ultralytics_YOLO_2023} performs poorly, but fine-tuning on even a small subset of our labeled dataset leads to large gains, especially in occlusion-heavy tasks like dressing and transfer. These results demonstrate the potential of \datasetname{} in advancing robust pose perception for real-world caregiving.}

\textbf{Occlusion}: Physical interactions in close proximity between caregivers and care recipients, along with the presence of assistive devices, result in frequent heavy occlusion. This makes pose estimation particularly challenging, \rev{due to partial visibility of the body}.\\
\indent\textbf{Distribution Shift for Real-world Caregiving}: \rev{The caregiving domain exhibits a substantial distribution shift compared to general human pose datasets.} It involves close physical interaction between two individuals, \rev{necessitating} multi-body pose estimation. \rev{Prior datasets}~\cite{sbu_kinect} \rev{typically} assume simpler poses (\rev{e.g.,} standing) and \rev{fixed camera placements} that minimize occlusion. Additionally, caregiving involves many unique postures that rarely occur in non-caregiving ADL \rev{scenarios}, further complicating pose estimation and generalization.
Finally, caregiving contains multimodal \rev{variability} in task planning for multiple tasks, much of which originates from OT expertise.
\rev{These factors pose challenges for existing models} for robot-assisted caregiving. Our dataset has the potential to bridge this gap.

\begin{table}[t]
\centering
\caption{2D Pose Detection Performance (mAP$_{50\text{--}95}$).}
\label{tab:2dpose_detection}
\small
\begin{tabular}{lccc}
\toprule
\makecell[c]{\textbf{Method} \\ \scriptsize (YOLOv11 Variant)} & \textbf{Bathing} & \textbf{Dressing} & \textbf{Transfer} \\
\midrule
Pretrained   & 0.0244 & 0.0259 & 0.0218 \\
Fine-tuned (1 OT)    & 0.5711 & 0.7052 & 0.5964 \\
Fine-tuned (5 OTs)   & \textbf{0.7757} & \textbf{0.8228} & \textbf{0.6648} \\
\bottomrule
\end{tabular}
\vspace{-0.2cm}
\end{table}

\begin{table}[t]
\centering
\caption{\small 3D Pose Estimation Performance (MPJPE in mm)}
\vspace{-0.1cm}
\label{tab:3dpose_detection}
\scriptsize
\setlength{\tabcolsep}{3pt} %
\begin{tabular}{lcccc}
\toprule
\textbf{Method} & RTMOPose3D~\cite{jiang2024rtmopose3d} & MixSTE~\cite{Zhang2022MixSTE} & HoT~\cite{li2024hot} & MHFormer~\cite{li2022mhformer} \\
\midrule
\textbf{MPJPE (mm)} & 119.9 & 122.9 & 142.2 & 162.7 \\
\bottomrule
\end{tabular}
\vspace{-0.7cm}
\end{table}

\vspace{-0.2cm}
\subsection{Planning: Long-horizon task recognition}~\label{Sec:Evaluation-planning}
We run VidChapters-7M~\cite{yang2023vidchapters} on a subset of 21 videos. \rev{Qualitative results are available on our website~\cite{website}.} While existing methods can recognize some subtasks, a significant gap remains in the caregiving domain. The lack of training data contributes to errors---for example, the model misidentifies ``positioning the Hoyer'' as ``positioning the foyer,'' \rev{likely due to unfamiliarity with caregiving terminology}. \rev{The long task horizons make it difficult for current models to recognize full procedures.} Finally, the diversity in task plans introduces challenges in \rev{decomposing} long-horizon task plans.  
\vspace{-0.1cm}

\section{Discussion}
\vspace{-0.1cm}
In this work, we proposed \datasetname{}, the first multi-task, \rev{multimodal}, expert-collected dataset for robot caregiving.
While the dataset is already large, future work could consider supplementing it with partial sensory data that is easier to obtain.
Another limitation of \datasetname{} is its focus on fully passive care recipients. Future work will consider partially mobile individuals who actively participate in ADLs.
These efforts are crucial to advance learning-based approaches, ultimately enabling more adaptable and capable robot caregivers.

\vspace{-0.2cm}
\bibliographystyle{IEEEtran}
\balance
\vspace{-0.2cm}
\bibliography{references}

\end{document}